\documentclass[review,OA]{AAAI-Std}
\usepackage{graphicx}
\def\artid{0001}%
\usepackage[T1]{fontenc}
\usepackage{bm}\usepackage{amsmath}

 \usepackage{xcolor}

\newcommand{\Xt}[1]{\bm{X}^{(#1)}}

\newcommand{\Yt}[1]{\bm{Y}^{(#1)}}

\def\x{\mathbf{x}}

\def\e{\mathbf{e}}

\def\S{\mathbb{S}}

\def\x{{\mathbf x}}

\usepackage{har2nat}
 \usepackage[utf8]{inputenc}
\begin{document}

\title{Robust Internal Representations for Domain Generalization}
 
{  \author{Mohammad Rostami}}

 \address{Mohammad Rostami is with  the University of Southern California. Contact: rostamim@usc.edu.}

\abstract[Abstract]{This   paper which is  part of the   New Faculty Highlights Invited Speaker Program of AAAI'23 \cite{rostami2023transfer}, serves as a comprehensive survey of my   research in   transfer learning by utilizing embedding spaces. The work reviewed in this paper specifically revolves around  the inherent challenges associated with continual learning and limited availability of labeled data. By providing an overview of my past and ongoing contributions, this paper aims to present a  holistic understanding of my research, paving the way for future explorations and advancements in the field. My research delves into the various settings of transfer learning, including, few-shot learning, zero-shot learning, continual learning, domain adaptation, and distributed learning.  I hope this survey provides a forward-looking perspective for researchers who would like to focus on similar research directions.}


\maketitle

\section{Introduction}

Advances in machine learning  (ML)  during the past decade have led to a significant performance improvement in a wide range of application with human-level performance being reached in many tasks. To a large extent, this success is a result of breakthroughs in  deep learning (DL). However, adopting DL in most applications relies on the availability of large annotated datasets for training deep neural networks. The emergence of crowdsourcing data labeling platforms like Amazon Turk has made data annotation feasible for general applications. These platforms allow to benefit from the wisdom of crowd by hiring individuals from diverse backgrounds to contribute to the data annotation process with minimal training procedures. We can often correct mislabels using aggregation techniques and reach expert-level annotation accuracies~\cite{rostami2018crowdsourcing}. However, generating annotation datasets remains a challenging, expensive, and complex task in many   applications. For instance, in the field of medicine, data annotation requires the expertise of trained physicians who are generally busy with vital tasks~\cite{avramidis2022automating,sun2023explainable}. Similarly, in non-natural image modalities such as synthetic aperture radar (SAR), data interpretation and annotation can only be effectively performed by trained experts who have years of experience~\cite{rostami2019sar}. When dealing with graph-structured data, human intuition would become weak which makes model interpretability more challenging~\cite{pope2019explainability,pope2018discovering}. Additionally, there are domains where data cannot be readily shared due to concerns related to security and privacy~\cite{stansecure}. As a result, crowdsourcing is not a viable solution for data annotation in these fields. The sensitive nature of the data and the associated risks in these applications necessitate alternative approaches to data annotation that ensure privacy and data confidentiality while making ML feasible.

Even if we are able to generate a high-quality annotated training dataset and then train a good model, its performance would degrade if we face distributional changes or drifts during the testing time after the initial training phase~\cite{rostami2021lifelong}. Distributional drifts lead to a distribution discrepancy on the testing dataset, necessitating expensive, time-consuming, and everlasting model retraining which in tern requires continual data annotations. Despite some behavioral similarities between deep neural networks and the nervous system~\cite{morgenstern2014properties},   the human nervous system can tackle these challenges considerably better. These similarities suggest that perhaps, we can equip deep learning with mechanism that the nervous system uses for improved performance. For example, Humans have the ability to learn many diverse problems efficiently in terms of training data points and, in some applications, much faster. In many cases, humans can learn a novel class by observing only a handful of examples and sometimes with no training instance, just by relying on high-level descriptions. Moreover, they adapt to changes fast and can generalize their knowledge to unexplored domains more straightforwardly. In contrast, current AI approaches primarily do not offer such abilities. These abilities seem feasible because humans do not start learning from scratch every time they learn and use acquired and accumulated knowledge from past experiences when learning new problems. Inspired by these abilities of natural intelligence, an important research direction in AI is to explore how an agent can store knowledge when it faces and learns multiple problems and then how it can use the acquired knowledge to improve learning quality/speed, and to overcome data scarcity in other related problems~\cite{rostami2019learning}. The hope is to broaden the applicability of ML to more domains and applications and develop intelligent agents that are more autonomous.

To improve AI, we need to go beyond the classic setting of learning a single problem in isolation. In my research, I explore the concept of learning not just   a single isolated machine learning problem, but   leveraging the relationships between multiple problems in order to enhance the learning process and avoid redundant learning. Traditionally, learning has been viewed as the search for a predictive function within the context of a single problem. However, this perspective fails to capture the potential benefits that can be gained from transferring knowledge between related problems.
We can consider a probabilistic framework, where the learning ability is acquired through an identically distributed dataset consisting of samples drawn from an unknown distribution. In this context, the aim for transfer learning is to exploit the connections between distributions of  different problems and utilize the acquired knowledge to improve learning outcomes across the problems. As a result, we leverage knowledge gained from one task to benefit another through updating the ML learning pipeline~\cite{rostami2021transfer}.

A prominent approach for knowledge transfer is to map data points originating from diverse distributions into a shared latent embedding space such that the acquired knowledge is more interpretable~\cite{rostami2021transfer}. In this shared space, the relationships and similarities between the distributions are encoded as geometric distances between the data representations. In other words, the embedding space represents data in terms of meaningful features that allow transferring knowledge across different problems. By doing so, knowledge can be effectively transferred across different distributions, encompassing tasks~\cite{rostami2020using}, domains~\cite{stan2021unsupervised},  and   agents~\cite{rostami2017multi,rostami2018lifelong}.  The intermediate data representations capture the similarities between the distributions in a higher level, allowing for the transfer of knowledge based on high-level similarities or descriptions~\cite{rostami2022transfer}. For example, two classification tasks that have different input spaces but both share the same classes, may have aligned distributions in the embedding space.
Geometric notions play a vital role in modeling similarities, enabling the formulation of coupled optimization problems that incorporate objective functions based on geometric distances of representations. 

The above approach allows for a problem-level analysis, where multiple probability distributions can capture the underlying relationships between the problems. For example, if the embedding space is designed to be task-agnostic when two problems share the same classes, a classifier trained on one problem can generalize well to other related problems due to sharing similar distributions in the embedding space. Additionally, by simultaneously learning several problems with a limited number of annotated data points, it becomes possible to leverage all available data across the domains effectively, leading to requiring less data to achieve a similar performance compare to single task learning or improve the learning speed~\cite{rostami2019deep, mirtaherione}. The shared embedding can also be used to relate the current task to past obtained experiences to learn it more efficiently, rather learning the task from scratch.
By considering the interplay between various problems and leveraging the relationships between them, we can enhance the learning speed, generalization, and overall performance of the learning process using less data. This approach extends the traditional isolated problem learning paradigm and opens up  possibilities for knowledge transfer and improved learning across diverse problems.

In my research, I examine the utilization of representation learning in embedding spaces as a means of transferring knowledge to address a set of significant machine learning problems. Specifically, I focus on the development of embedding spaces that effectively capture and represent data in a desired format.
Through my investigation, I found that despite variations in the transfer learning scenarios, such as the direction of knowledge transfer and the organization of data, they can benefit from this common perspective. My contributions to the field encompass two main areas of ``continual learning'' and ``learning with limited labeled data'' which will be survey in what follows.

\section{Framework Formulation} 

 In our framework, we consider more than one ML problem to solve.
Consider a set of countable ML problem  $\mathcal{T}^{(u)}, u\in \mathbb{U}$, where $\mathbb{U}$ is a countable set. Each task is characterized by an unknown distribution $p^{(u)}(\bm{x},\bm{y})$ defined over the  vector space $\mathcal{X}^{(u)}\times \mathcal{Y}^{(u)}$. Learning each problem in this set can be considered  solving for a predictive function $f^{(u)}:\mathcal{X}^{(u)}\rightarrow \mathcal{Y}^{(u)}$ that minimizes the expected error during testing. Since the data distribution is unknown, we opt for minimizing the empirical error on a given training dataset $\mathcal{D}^{(u)}$ to solve for an optimal model. If we select a complex enough family of predictive functions and have access to a large enough annotated training dataset, the empirical error would be a good surrogate for the expected error. 

Rather learning these problems in isolation, we consider learning them by relating them, depending on the specific learning setting, to improve the collective performance. For this purpose, we assume that the predictive functions $f^{(u)}(\cdot)$ can be written as the combination of two functions $f^{(u)}(\cdot)= h^{(u)}\circ \psi^{(u)}(\cdot)$, where $\psi^{(u)}(\cdot):\mathcal{X}^{(u)}\rightarrow \mathcal{Z}$ is an encoder and $h(\cdot):\mathcal{Z} \rightarrow \mathcal{Y}^{(u)}$ is a secondary classifier that receives its input from the encoder. In this formulation, $\mathcal{Z}$ denotes the shared embedding space which is an intermediate space between the input and the the output spaces. The embedding space couples the problems and allows for relating the problems. The broad idea that we want to explore is to train the mapping functions such that the relations between the problem emerge as interpretable features in the latent space. We optimize the following objective function to learn the set of problems:
\begin{equation}
\begin{split}
&\min_{ f^{(1)},\ldots,f^{(u)}}\sum_{u=1}^U \lambda^{(u)}\mathcal{L}^{(u)}(f^{(u)}(\Xt{u}) )+\\& \sum_{u,v=1}^U \gamma^{(u,v)}\mathcal{M}^{u,v}(\psi^{(u)}(\Xt{u}), \psi^{(v)}(\Xt{v}))
\enspace,
\end{split}
\label{eq:maineqforSurvey}
\end{equation} 
where $\mathcal{L}^{(u)}(\cdot)$ denotes a single problem-level  loss function, e.g., empirical risk using the cross-entropy loss on an annotated dataset or a regularization term on the model parameters. The function level operator
$\mathcal{M}(\cdot,\cdot)$   denotes measures of  similarity or relations between any two problems, and  $\lambda^{(u)}$ and $\gamma^{(u,v)}$ are two sets of trade-off parameters to tune the importance of the loss terms.
In Equation~\ref{eq:maineqforSurvey}, the first additive term denotes the problem-level loss functions that depend on a single domain. The second sum   encodes prior knowledge about the pair-wise similarities between the problems and through coupling the problems, enforces knowledge transfer through the shared embedding space across the tasks. 

Equation~\ref{eq:maineqforSurvey} formulates many different learning settings, including, zeros-shot leaning, few-shot learning, domain adaptation, and continual learning, based on the specific way that we define the loss terms. In the next sections, we will see how the accurate definitions of these terms, instantiates Equation~\ref{eq:maineqforSurvey} as the ultimate problem we solve in each problem setting.   I have primarily focused on  addressing the challenges of two broad areas of ``data scarcity'' and ``learning in the presence of temporal drifts in the input space''.
In the  first domain, the goal is to benefit from a data-rich auxiliary problem to solve a data-scarce problem. In the second domain, the goal is to improve performance on several tasks using knowledge transfer.

\section{Learning in the Presence of Label Scarcity}

Label scarcity can manifest in various learning scenarios, necessitating the use of alternative sources of information to overcome the challenge of learning from unlabeled data. One such scenario is low-shot learning, which includes few-shot learning and zero-shot learning. In these settings, we usually have two distinct sets of classes. The first set comprises of high-resource classes for which an ample number of annotated samples are available. Conversely, the second set consists of low-resource classes with only a few or no labeled samples. The objective in these settings is to establish relationships between the low-resource and high-resource classes through an auxiliary domain, leveraging the knowledge gained from the high-resource classes to learn effectively in the low-resource setting. In  contrast, unsupervised domain adaptation presents a different learning scenario where labeled samples are unavailable for all classes. Instead, the approach involves leveraging a secondary target domain that possesses labeled data and shares strong relationships and similarities with the source domain. For example, both domains may share exactly the same classes. By leveraging the labeled samples from the target domain, the goal is to bridge the domain gap and facilitate knowledge transfer from the labeled target domain to the unlabeled source domain.

Within these distinct learning settings, characterized by the challenge of label scarcity, I have undertaken the task of addressing the associated challenges by employing shared embedding spaces. In the context of low-shot learning, my research has been dedicated to developing methodologies that facilitate the establishment of effective relationships between low and high-resource classes through the shared embedding space. The latent space  captures semantic meaning, enabling the transfer of knowledge and facilitating robust performance on the low-resource classes. By leveraging the shared embedding space, I aim to bridge the gap between limited labeled data and the need for accurate classification and recognition in low-shot learning scenarios.
Furthermore, my focus extends to the field of unsupervised domain adaptation, where I have delved into various techniques and algorithms. The objective in unsupervised domain adaptation is to address the disparity between an unlabeled target domain and a labeled source domain. In this problem, the goal is to narrow the domain gap, allowing for effective knowledge transfer from the labeled source domain to the unlabeled target domain. The solution involves developing methods to align the distributions of the source and target domains, ensuring that the model can generalize well to the target domain despite the lack of labeled data. Through the application of   domain adaptation techniques, we can improve the performance and adaptability of machine learning models in real-world scenarios where labeled data is scarce or unavailable in the target domain.

\subsection{Zero-Shot Learning}

The objective of zero-shot learning (ZSL) is to enable a model to make generalizations about a class or a task even when it has no training data specifically related to that class or task. As expected, it is not feasible to learn without any training data. However, learning becomes possible if we can leverage additional information from a secondary source of information for which data is available.
One instance of ZSL is for multi-class classification, where the goal is to learn and classify a set of unseen classes based on the knowledge gained from learning seen classes. This approach is   useful for learning classes that occur infrequently or emerge after the initial training phase.
The key to bridging the gap between seen and unseen classes lies in the domain of semantic textual descriptions. Nowadays, we have   access to textual descriptions of various classes through public encyclopedias like Wikipedia. These textual descriptions serve as a valuable resource for understanding and connecting the seen and unseen classes and transfer what we learn from learning the seen classes to the unseen classes.
By utilizing the information contained in these descriptions, ZSL algorithms can extract meaningful features and attributes of both seen and unseen classes. This possibility enables the model to generalize its knowledge, gained from the seen classes, to the unseen ones, even in the absence of annotate training data for the unseen classes. In essence, the semantic textual descriptions act as a link that facilitates  knowledge transfer between different classes, allowing the model to expand its capabilities beyond the initially trained classes.

To tackle ZSL, we  train mappings that allow us to transfer data from the visual domain to the semantic domain. This task is achieved by leveraging a set of seen classes for which we have accessible training data in both modalities. If we can successfully learn such a mapping, we can map an image from unseen classes to its corresponding textual description and identify its class in in the semantic domain. Let's denote the visual domain as $\mathcal{X}^{(v)}$ and the textual domain as $\mathcal{X}^{(t)}$.
In order to train our mapping, we utilize a training dataset $\mathcal{D}$ consisting of visual data $\Xt{v}$, textual data $\Xt{t}$, and visual labels $\Yt{v}$ for the seen classes. 
Additionally, we have access to a separate textual dataset $\mathcal{D}'$ specifically for the unseen classes. This dataset consists of textual data $\Xt{t'}$  and  visual labels $\Yt{v'}$.
Now, we specialize Equation~\ref{eq:maineqforSurvey} which pertains to ZSL:
\begin{equation}
{
\min_{ \theta^{(v)},\theta^{(t)}}\sum_i \ell(\psi^{(t)}_{\theta^{(t)}} (\bm{x}_i^{(t)}),\psi^{(v)}_{\theta^{(v)}} (\bm{x}_i^{(v)}))
\enspace,
\label{eq:ZSLSocherSurvey}
}
\end{equation} 
In this context, the parameters $\theta^{(v)}$ and $\theta^{(t)}$ are model parameters that need to be learned and $\ell$ represents a point-wise similarity metric. Essentially, when we solve Equation \ref{eq:ZSLSocherSurvey}, we aim to learn the mappings $\psi^{(v)}$ and $\psi^{(t)}$ to ensure that the textual and visual information for seen images are embedded closely together in the shared embedding space.
Once we   solve Equation \ref{eq:ZSLSocherSurvey}, we can utilize the learned mappings for ZSL of unseen classes. We employ the $\theta^{(t)}$ mapping to map the semantic descriptions of all unseen classes into the shared embedding space. Additionally, we utilize $\theta^{(v)}$ to map images from unseen classes into the shared embedding space.
By performing these mappings, ZSL becomes feasible as we can now search for the class description that is closest to an image from an unseen class. Hence, we can determine the most appropriate class label for an image even if it belongs to a class that has not been encountered during the training phase through matching the image to the most relevant description.

There are many ZSL methods that solve Equation \ref{eq:ZSLSocherSurvey} through choosing various options for the similarity metric and the mappings. Building upon prior my works to coupled two domains in the area of sparse representations and incorporating side information in improving the sparse recovery problem
\cite{rostami2012image,rostami2013compressedsid,rostami2013compressed,rostami2012gradient,rostami2015surface,hashemi2016efficient}, we used the idea of coupled dictionary learning \cite{rostami2011image,rehman2012ssim} for ZSL.
Our idea is to use the seen classes to train two coupled dictionaries $\bm{D}$ and $\bm{L}$ such that for all images from the seen classes, we have $\bm{x}_i^{(t)}= \bm{D}\alpha_i^{(t)}$ and $\bm{x}_i^{(v)}= \bm{L}\alpha_i^{(t)}$, where $\alpha_i^{(t)}$ is trained to be a sparse vector~\cite{kolouri2018joint,kolouri2021attribute,rostami2022zero}.
After training the coupled dictionaries, we can use sparse recovery algorithms such as LASSO to solve for the sparse vectors that correspond to semantic descriptions and images  and then use Euclidean distance to match the appropriate semantic description to a given image. Moreover, we also mitigated domain shift problem in ZSL using   entropy regularization. Domain shift happens because the base model is biased towards predicting the seen classes.  We also used label propagation to address  hubness in higher dimensions because simply using Euclidean distance to search for the right label faces the challenges posed by the curse of dimensionality.

We also extended our idea to ZSL of reinforcement learning (RL) tasks \cite{isele2016using}. In this setting, our goal was to enable learning RL tasks through high-level descriptions of a given task.
Our approach involves transferring knowledge between several RL tasks that are similar, each described using the high-level descriptors that have different numeric values to instantiate each task. Our core idea is to learn a mapping that maps the high-level descriptors to the optimal policy that is   learned using policy gradient methods.
To address this challenge, we extended our approach to learn such a mapping based on coupled dictionary learning \cite{isele2016using}. Our method leverages high-level task descriptions to model the inter-task relationships between the set of tasks that are defined using the same types of descriptors that can vary in value. By utilizing these task descriptors, we reduce the burden of relying on task-specific training data for estimating the relationships between tasks.

In our work, we demonstrate that incorporating task descriptors significantly improves the performance of learned task policies because high-level descriptors can be leveraged as a secondary source of information  \cite{isele2016using}. This improvement is supported by both theoretical justifications and empirical evidence across various RL problems. By effectively modeling the inter-task relationships using task descriptors, our approach enhances the efficiency and effectiveness of RL because past experiences can be used to not learning the task at hand from scratch.
When only the descriptor for a new task is accessible, we can accurately predict a model for the new task, enabling ZSL using the trained coupled dictionaries in a process very similar to ZSL in the visual classification setting \cite{isele2016using}. We eliminate the need to collect training data for RL specifically for addressing the new task, which is extremely helpful in RL settings. By leveraging the learned inter-task relationships, our approach enables generalizing knowledge to unseen tasks and efficiently tackles new tasks without extensive data collection.
We also showed that our work can address ZSL for regression and classification tasks \cite{rostami2020using}, as far as these tasks can be defined by a set of high-level task descriptors.

\subsection{Few-Shot Learning}

Although few-shot learning (FSL) and zero-shot learning (ZSL) seem to be very similar, they have distinct characteristics and objectives.
 Within a multi-class classification setting, the goal in FSL is to enable the model to generalize and make accurate predictions on a number of classes with limited training samples. Hence, FSL serves as a bridge   between classic supervised learning and ZSL.
In contrast to ZSL, we usually do not rely on semantic descriptions as an auxiliary domain to address the challenges of FSL. The primary solution is still to benefit from a set of high-resource classes and transfer learning to enable FSL of low-resource classes but primarily using solely the visual domain. A primary approach in FSL is to compute class prototypes in an embedding space using the few training examples to represent the class well. Classification then can be preformed by computing the similarity between a query and these prototypes in terms of a suitable metric. We can either use predefined metrics to measure similarity or train a metric for this purpose.  I have also explored FSL in    domain adaptation \cite{rostami2019sar}.

In the context of FSL for object recognition, the presence of unlabeled instances from novel classes that are not included in the fixed set of the training classes but appear in the background can introduce label noise, leading to a decline in FSL performance. The reason is that these objects are going to be classified as one of the classes in the training set, despite being relatively different. In our recent work \cite{shangguan2023identification,shangguan2023improved}, we proposed a semi-supervised algorithm that addresses this issue by detecting and leveraging these unlabeled novel objects as positive samples during the training process and assigning them to a new background class using contrastive learning. As a result, we enhance FSL performance by differentiating novel objects from the base training classes.
To identify the novel classes, we introduced a hierarchical ternary classification region proposal network (HTRPN) that effectively localizes potential unlabeled novel objects present in the background and assigns them new objectness labels. By incorporating a hierarchical sampling strategy within the region proposal network (RPN) of the base Faster RCNN model, we enhance the object detection  capability to perceive and accurately detect larger objects.
Through our approach, we successfully tackle the challenge posed by unlabeled novel objects, leading to noticeable improvements in FSL performance. By effectively utilizing these unlabeled instances and improving the detection process, our method enhances the effectiveness of FSL for object recognition tasks by reducing interference caused by the novel objects.

  I have also focused on studying FSL within the context of relational learning on temporal knowledge graphs (TKGs) \cite{mirtaherione}.
A knowledge graph (KG) is a structured representation that captures the relationships between a collection of entities.
In various real-world applications, KGs possess a long-tailed frequency distribution for their relationship types, with the majority of relations being infrequent. Consequently, it is   desirable to develop FSL methods that can generalize from only a few examples for these rare relations. We utilize a self-attention mechanism to effectively capture the temporal interactions between entities present in a TKG. Specifically, we introduce a temporal neighborhood encoder that incorporates the self-attention mechanism to efficiently  extract the temporal  neighborhood information pertaining to each entity. By encoding these interactions, we are able to calculate a similarity score between a given query and a one-shot example, enabling accurate predictions even when training data is limited based on similarities with the exiting relations in the TKG. Notably, our approach demonstrates significant performance improvements    specifically for sparse relations, addressing a common challenge encountered in real-world TKGs.

\subsection{Unsupervised Domain Adaptation}

The goal in an unsupervised domain adaptation (UDA) setting is to transfer knowledge from a source domain, where labeled data is accessible, to a distinct target domain, where labeled data is scarce. As a result, we relax the need for data annotation in the target domain. UDA is feasible if the two domains share similarities and annotating data in the source domain is simpler. The classic assumption for similarity of works in UDA is that both domains share the same classes. 
The classic UDA setting can model practical setting, where the two domains differ in  aspects such as imaging conditions, sensor types, data collection protocols, or data generation. Another important application is using controlled synthetic data to train models for real data. For example, the source domain may be the domain of natural images and the target domain may be the domain of synthetic aperture radar (SAR) images. In other words, the two domain usually correspond to distinct data modalities or types.  The differences between the two domains would lead to domain shifts which can lead to a degradation in performance when applying a model trained on the source domain to a target domain. When the two domains share the same classes, the naive solution is to use the source-trained model on the target domain but we aim to adapt this baseline model  to perform better on the target domain.

A broad approach to address UDA is to map data points from both domains into a shared embedding space, where domain-invariant representations or features   capture the underlying shared similarities between the two domains. By aligning the statistical properties of the features for the source and target domains in the shared domain-agnostic embedding space, UDA methods aim to reduce the distribution discrepancy and enable generalization of a  classifier that receives its input from the shared embedding space and is trained using the source domain annotated data  on the target domain. To this end, we can model these mappings using two deep neural network encoders which are trained to align the two distributions at their shared output spaces.
More specifically, we can use Equation~\ref{eq:maineqforSurvey} to formulate the classic UDA problem as solving the following domain alignment problem:
\begin{equation}
\begin{split}
\min_{\theta^{(s)} ,\theta^{(t)} ,\kappa^{(t)} }& \mathcal{L}^{(s)}(h^{(s)}_{\kappa^{(s)}}(\psi^{(s)}_{\theta^{(s)}}(\Xt{s})),\Yt{s} )+ \\&\gamma \mathcal{M}(\psi^{(s)}_{\theta^{(s)}} (\Xt{s}), \psi^{(t)}_{\theta^{(t)} }(\Xt{t}))
\enspace,
\end{split}
\label{eq:UDAeqforSurvey}
\end{equation} 
where $\theta^{(s)}$ denotes the source domain encoder parameters, $ \theta^{(t)}$ denotes the target domain encoder parameters, and $\kappa^{(t)}$ represent the shared classifier parameters. In Equation~\eqref{eq:UDAeqforSurvey}, the first term corresponds to the empirical error for the source domain which is computed using annotated data. The second term is a cross-domain alignment term that encourages the two domains to have similar distributions. By solving this optimization problem for the optimal functions $\psi^{(t)}, \psi^{(s)},$ and $h^{(s)}$, the data representations in the shared embedding space at the output of the encoders become both discriminative for classification and at the same time invariant to domain differences. Consequently, the source-trained classifier  $h^{(s)}$ will generalize on the target domain distribution. Depending on the amount of domain gap, sometimes we can let $\psi^{(t)}= \psi^{(s)}$ and use the same shared encoder for domain alignment and solve a less complex problem.

There are two primary approaches for domain alignment in UDA.
  A prevalent approach  is based on adversarial learning~\cite{jian2023unsupervised,jian2023unsupervisedj,murez2023domain,gabourie2021system}. Within the context of adversarial learning, the domain discriminator network is trained to differentiate between samples from the source and target domains.  On the other hand, the generator network aims to confuse the domain discriminator by learning domain-invariant representations. By optimizing the generator network, the objective is to minimize the ability of the discriminator to distinguish between the source and target domains. Consequently, the embedding space at the output of the generator network becomes domain-agnostic when the domain discriminator network fails to differentiate between the two domains effectively.
Another primary technique for UDA involves minimizing the discrepancy between the source and target domain distributions directly. This approach focuses on aligning the two distributions to facilitate better  knowledge transfer. Two popular methods for measuring the discrepancy are maximum mean discrepancy (MMD) and the optimal transport metric. These methods aim to find optimal mappings or alignments between the source and target distributions by minimizing the selected  discrepancy measure between the two distributions. More recent methods benefit from secondary mechanisms within each primary approach to improve model generalization. In my research, I have primarily explored the latter approach for domain alignment.

In our initial work for UDA, we focused on addressing the challenge of aligning the source and target domains directly. To achieve this goal, we employed the sliced Wasserstein distance (SWD) in Equation \ref{eq:UDAeqforSurvey}  to minimize the discrepancy between the probability distributions of the source and target visual domains \cite{gabourie2019learning}. 
However, aligning the distributions may not lead to an optimal performance. Another challenge arises in aligning the classes between the source and target domains consistently. To overcome this challenge, we utilized high confidence pseudo-labels for the target domain. These pseudo-labels were assigned based on the highest prediction   obtained from the classifier trained on the source domain.
By incorporating pseudo-labels, we extended Equation \ref{eq:UDAeqforSurvey} to align the two distributions in a class-conditioned manner. As a result, we can match samples that shared the same class across the source and target domains. The objective was to establish correspondence between classes in the embedding space and avoid matching inconsistent classes across the two domains. This class matching strategy increased the performance across several UDA benchmarks and led to competitive performance against several existing approaches. 
Subsequently, we improved upon this baseline by increasing the interclass margins in the embedding space \cite{rostami2022increasing,rostami2022increasingj,rostami2020learning}.
We also provided a theoretical analysis to demonstrate that our approach minimizes an upperbound for the expected error on the target domain.

In order to increase the margins between different classes \cite{rostami2022increasing}, we employ a method that estimates the internally learned multi-modal distribution within the source domain, where each distribution mode corresponds to one of the input classes. This distribution is obtained during the pretraining phase on the source domain and can be estimated using a Gaussian mixture model (GMM). By estimating this distribution, we aim to enhance the separation between classes in the source domain, thereby mitigating the impact of domain shift in the target domain.
To achieve this goal, we utilize the GMM to generate random samples that are close to the mean for each component of the GMM. These samples effectively represent the different modes within the distribution. We leverage these samples to push the representations of the source domain data representations away from the boundaries between the distribution modes through minimizing the distance between the empirical distributions of the GMM   and the source domain distribution in the embedding space. This process helps in making the extracted features more compact, leading to improved interclass separation.
We demonstrate the enhanced generalizability of models across various benchmark datasets used for UDA in image classification \cite{rostami2022increasing}. We also demonstrated that conditioned on selecting the appropriate encoders, our approach can be used to address UDA for classification of sentiment analysis \cite{rostami2021domain} and graph-structured data \cite{wu2023unsupervised}.

We extended our approach to address the challenge of data annotation in SAR domain problems \cite{rostami2019deep,rostami2019sar,kolouri2020systems}. In contrast to the EO (Electro-Optical) domain, labeling data in the Synthetic Aperture Radar (SAR) domain poses greater challenges~\cite{rostami2022system}. Due to various reasons, including security regulations and necessity of extensive training to interpret SAR data, utilizing crowdsourcing platforms for labeling SAR data is challenging.  
Our approach is to reduce the number annotated data in the SAR domain based on knowledge transfer from a related problem in the EO domain, where obtaining labeled data is comparatively easier.  To accomplish this goal, we trained two deep encoders that share the same last layer to represent the shared embedding space. These encoders map data points from both the EO and SAR domains to the shared embedding space and then are trained to align the two distribution by minimizing the SWD metric. We use distinct encoders because the domain gap between the EO and SAR domains are more significant. Additionally, we leverage a limited number of labeled SAR data points to match the distributions in a class-conditioned manner to improve performance.
Experiments using real-world datasets demonstrated that our method is effective.

In classic UDA, the goal is to adapt a model trained on labeled data from a source domain to perform well on an unlabeled target domain when data for both domains is accessible at the same time. 
Source-free domain adaptation refers to a setting where the adaptation process is performed without accessing any labeled data from the source domain during the adaptation phase.  This situation often arises when the source domain data is unavailable due to the sequential nature of the problem or comes with privacy constraints.
Hence, the focus shifts towards relaxing the need to process the source data directly and finding ways to adapt models without accessing the labeled source data during the adaptation phase.
Although source-free domain adaptation presents additional challenges due to the lack of labeled source data, it also presents opportunities for the development of innovative algorithms and techniques that can adapt source-trained models to new domains using only target domain data. By addressing this problem, we can extend the applicability of domain adaptation methods to a wider range of scenarios and applications.

To tackle the source-free UDA problem, we have developed an algorithm based on estimating the internal distribution as a GMM \cite{rostami2023overcoming,rostami2023overcomingj}. The estimated GMM   serves as a surrogate for the source domain distribution in Equation~\ref{eq:UDAeqforSurvey}. By aligning the target domain distribution with the estimated GMM, we enable effective knowledge transfer from the source domain to the target domain without direct access to the source domain data.
Furthermore, we have extended our approach to address UDA in semantic segmentation tasks for both natural images \cite{stan2021unsupervised,stan2022unsupervised} and medical domain images \cite{stan2022domainBMVC}. Our approach reaches performance levels comparable to classic UDA algorithms and is specifically helpful in medical domains, where sharing data is not highly regulated. Our results demonstrates the versatility and applicability of our approach across different domains and problem contexts.
Additionally, we have expanded our framework to handle UDA with multiple source domains \cite{stansecuremulti}, where we consider the presence of multiple source domains with private data. By incorporating multiple source domains, we leverage the diversity and complementary information across the sources to enhance the adaptation process and the performance on the target domain without sharing data.
Through these advancements, our approach contributes to the field of source-free UDA by providing effective solutions for adapting models in scenarios where direct access to the source data is not feasible or desirable.

\section{Continual Learning}
The second primary field in my research is
continual learning (CL). The primary objective in
CL, also referred to as lifelong machine learning (LML), is  to equip AI systems with the ability to continuously acquire and retain knowledge from an ongoing stream of data over an extended period and accumulative learn and improve themselves to handle new situations. Unlike classic ML that relies on static datasets and one-shot training-testing procedures, CL aims to develop techniques that enable AI systems to adapt and learn continuously as new data becomes available, without requiring a complete retraining process from scratch.
CL has immense potential and applicability in various domains, including robotics, autonomous vehicles, and reinforcement learning when the goal is to build autonomous intelligent systems that can improve themselves with minimal human intervention and leverage their past experiences to handle unforeseen situations effectively.
The significance of CL lies in its ability to tackle real-world scenarios where data distribution is non-stationary or evolves over time and an AI agent interacts with such an environment. Instead of treating each task   in isolation, CL takes into account the temporal nature of the data, allowing AI systems to adapt and incorporate new knowledge while preserving previously learned information when encountering new situations. By continually updating their   skills, CL agents can better cope with dynamic environments and effectively handle novel tasks or situations that were not encountered during initial training.
To achieve successful continual learning, various approaches and techniques have been developed in the literature.  

We need to address two key challenges CL. The first challenge, known as ``catastrophic forgetting'', arises when a model tends to forget previously acquired knowledge while learning new tasks. This phenomenon occurs because the model's parameters are updated to accommodate new tasks, leading to deviations from the optimal values that were once conducive to past learned tasks. As a result, the model's performance on previously learned tasks may significantly deteriorate.
To mitigate catastrophic forgetting, CL methods employ various primary techniques. One common approach is model regularization, where regularization mechanisms are introduced to stabilize the model's parameters and prevent drastic changes that could undermine past learned knowledge. By constraining the parameter updates, the model can strike a balance between accommodating new tasks and retaining valuable information from previous tasks.
Another widely used technique to address catastrophic forgetting is experience replay. This approach involves storing and replaying samples from past tasks during the learning process of new tasks. By revisiting and interleaving past experiences, the model has the opportunity to reinforce the knowledge learned in earlier tasks. This strategy helps mitigate forgetting by providing the model with a diverse range of training samples that span both old and new tasks.

Most existing works in CL focus on addressing  catastrophic forgetting.
But in addition to addressing catastrophic forgetting, another challenge in CL is leveraging past experiences to facilitate more efficient learning of current tasks through transfer learning. Transfer learning allows the model to transfer knowledge and representations learned from previous tasks to accelerate the learning process for new tasks. By utilizing relevant information acquired from prior experiences, the model can bootstrap its learning, potentially reducing the amount of training data required and improving generalization performance on the current task.
Both model regularization and experience replay can facilitate transfer learning.
Model regularization use a notion of parameter isolation, where each task is learned through distinct parameters and transfer learning is realized through the shared parameters. Architectural modifications is a similar idea, where new parameters are added to learn new tasks through these parameters. As a result, transfer learning is realized through shared parameters across similar tasks. To implement experience replay, we store a representative subset of training datasets for each past learned tasks in a memory buffer and then replay them back along with the new task data when a model is updated to learn a new task. These data points are selected such that they are informative and are a good representative of each training dataset. Transfer learning becomes feasible because the model has learned past tasks when it is updated. As a result, we do not initialize the model with random weights and start learning from  favorable initial   values that contribute to transferring knowledge.

In my research, I have focused on addressing catastrophic forgetting by using generative replay. One limitation of using a fixed-size memory buffer for experience replay is that the samples in the buffer becomes less representative of the pas tasks as the agent learns more tasks. As the number of tasks increases, there is limited space in the memory buffer, forcing the agent to discard samples from past tasks to make room for new task samples. As a result, the per task budget that we have for storing samples, reduces as the number of tasks grow, making the samples in the memory buffer less representative.

Drawing inspiration from the principles of complementary learning systems theory in neuroscience, I have developed an approach that incorporates a decoder into a base classifier, creating an autoencoder \cite{rostami2019complementary}. The decoder is added from the last year of the the base classifier with a mirror architecture such that the encoder and the decoder form an autoencoder. The created autoencoder is capable of generating pseudo-data points that resemble the true data points from past tasks. By operating at an abstract level, we capture the essential characteristics of past experiences and utilize them to enhance the learning process.
To generate pseudo-data points, the idea is to estimate the distribution that is formed in the autoencoder bottleneck layer. We then use this distribution to draw random samples and then feed them to the decoder network to generate pseudo-data points.
To estimate the learned distribution in the embedding space, we employ a GMM estimation which serves as the basis for generating data points that encapsulate the knowledge from past tasks. By sampling from the GMM and incorporating the generated samples into the experience replay process, we mitigate the detrimental effects of catastrophic forgetting.
Our framework is designed to update the model in a manner that ensures the shared distribution in the embedding space is relevant and beneficial for all tasks. When updating the model during the learning process, we specifically aim to align the internal distribution of the current task with the estimated GMM obtained from the previous time step.
By coupling the current task internal distribution and the past experiences encoded in the GMM, the model retains and utilizes the valuable knowledge gained from previously learned tasks, enabling it to adapt and generalize well to new tasks. This coupling of the current task with the past experience not only mitigates the negative effects of catastrophic forgetting but also facilitates transfer learning.
The model is able to build upon its existing knowledge, learn the current task efficiently, and at the same time mitigate forgetting effects via the generated samples.  

We have also expanded our framework to accommodate a setting where the tasks that come after the initial task are unlabeled  \cite{rostami2020generative,kolouri2021systems}. The motivation behind this setting stems from the challenge of concept shift, which assumes that the same learned concepts may undergo a distributional shift over an extended period of time and as a result  the statistical properties of the target classes change. These changes in turn invalidate the model on the drifted versions of the concepts.
To address the concept shift challenge, building upon the core ideas of our framework, we update the model such that the internal distribution remains stable as we learn the concepts in new domains. In other words, we consider each domain to encompass a period during which the concepts can be considered non-stationery. Because drifts usually happen smoothly, the distributions between the tasks have overlapping supports which makes our method extendable. As a result of our framework, the same learned concepts are expanded and adapted as we encounter new domains or environments.
Generative replay allows us to maintain our knowledge about the previous forms of the concepts using experience replay.  In short, our continual concept learning framework enables the model to  retain its previously acquired knowledge and   expand and refine the base concepts as it encounters new situations.

In a subsequent work, we have developed an algorithm to enable deep neural networks to learn new concepts and expand their knowledge incrementally in a CL setting \cite{rostami2021cognitively,rostami2023system}. Our approach is based on extending the idea of generative experience replay to an incremental setting.
Central to our algorithm is the integration of the parallel distributed processing theory, which aligns well with our approach. According to this theory, abstract concepts can be effectively encoded and represented in an embedding space, which takes the form of a multimodal distribution. By mapping these concepts to the embedding space, which is modeled by the bottleneck layer of an autoencoder, we are able to capture their fundamental characteristics and facilitate incremental learning.
One key assumption in our algorithm is that as we learn new tasks and encounter new concepts, the number of modes in the internal distribution of the embedding space increases to accommodate these novel classes. Simultaneously, we ensure that the modes representing previously learned concepts are aligned with the modes of these concepts present in the current task. As a result, our model is capable of learning new concepts while retaining knowledge about previously learned ones.
By incorporating this approach into our algorithm, we enable learning and expanding  knowledge about continually emerging concepts in an incremental learning scenario. The model dynamically allocates new modes to represent novel concepts while keeping the existing modes stable, allowing for seamless integration of new information while preserving the knowledge acquired from past tasks.  In what follows, I describe works in my research that expand continual learning ability  beyond the classic CL setting for various applications and   recent models such as transformers.

As face photos are becoming the prevalent biometric modality,  antispoofing against face spoofing attacks in face recognition systems is becoming extremely important. Spoofing attacks involve presenting fake or manipulated biometric information, such as photographs to gain unauthorized access to  a system. We developed a face antispoofing systems   when dealing with novel types of attacks that were not encountered during training phase \cite{rostami2021detection}. Most existing algorithms for this purpose, assume that all attack types are know a priori. Our algorithm enables continual detection of new attack types and self-adaptation to identify new spoofing types after the initial detection phase. We first enable  identifying novel attack types as anomalies  observed in the input data stream. We achieve  this goal by suppressing the confidence level of the network when the input data falls outside the internally learned distribution for the training samples. By doing so, we provide the network with the ability to detect and flag instances that deviate from the learned patterns, thereby identifying potential new attack types.
To ensure the network's adaptability to newly encountered attacks in a continual learning setting, we employ experience replay. Our algorithm allows us to update the model by incorporating knowledge about these new attack types without causing it to forget the previously learned attack types. By replaying past experiences during the training process, we reinforce the network's understanding of known attack patterns while simultaneously enabling it to learn and adapt to new patterns.
The experimental results on two real-world face spoofing datasets demonstrate the efficacy of our method in   detecting both known and unknown attack types, showcasing its potential in enhancing the performance of face antispoofing systems.

Most works on UDA only consider a single source and a single target domain. However, we can consider settings that new domains are encountered sequentially in a UDA setting.
For this purpose, we formulated a UDA problem in a continual learning context \cite{rostami2021lifelongNIPS}. Our objective  is to continually update a model in order to adapt to distributional shifts that occur as new tasks arrive sequentially, when labeled data is not available for these tasks. The formulation is different from classic CL because   CL algorithms primarily focus on handling tasks that come with labeled data. We   bridge the gap between UDA and CL settings without the need for labeled data using a two-step approach. Firstly, we leverage the learned internal distribution of the model to enhance its generalization capability on new domains. By consolidating the knowledge gained from past tasks and incorporating it into the model's internal distribution, we enable the model to adapt and generalize effectively to new domains. Secondly, we address catastrophic forgetting based on experience replay using a memory buffer. Using a memory buffer allows addressing tasks with more complex data compared to generative replay because data generation is a more challenging task compared to storing a subset of samples. To select the representative samples, we rely on the learned internal distribution. We select the samples that lie close to the modes for each class.

As AI is being adopted increasingly for decision-making processes, one of the major areas of concern in the AI community is existing biases in these algorithms.
Current fairness algorithms   typically employ a single-shot training approach, where an AI model is trained on a labeled dataset containing sensitive attributes such that bias is minimized on the dataset. While this training strategy works well when dealing with problems that have stationary distributions,   it becomes vulnerable to distributional shifts in the input space that may occur after the initial training phase. We develop an algorithm
to address this issue  that enables a fair model to maintain fairness even under domain shifts, using only new unlabeled data points ~\cite{stan2023preserving}. Building upon the previous works we did, we formulate this learning scenario as a UDA problem in a CL setting. Our algorithm focuses on updating the model in a way that the internal representation of the data remains unbiased, despite distributional shifts in the input space. 
To validate the effectiveness of our algorithm, we conducted extensive empirical experiments on three well-known fairness datasets by first demonstrating that bias under domain shift is a real-world concern.  We demonstrated that our algorithm successfully preserves fairness in the face of domain shifts, even when working with new unlabeled data points.

We have also explore the problem of TKG completion in a CL setting \cite{mirtaheri2023history}.
In real-world scenarios, TKG data is typically received gradually as events unfold, leading to a dynamic and ever-changing distribution of data over time. While it is possible to incorporate fine-tuning into existing methods to adapt them to the evolving TKG data, this approach often leads to forgetting effects on past events. Essentially, the model tends to lose previously learned patterns and knowledge when it focuses too much on new information. On the other hand, starting the training process from scratch can address the issue of forgetting, but it is computationally intensive and not feasible in many cases.
To address these challenges, we formulated the TKG completion problem CL problem. Our framework builds upon two key concepts: temporal regularization and clustering-based experience replay.
Firstly, temporal regularization is used to encourage the model to repurpose its less important parameters for learning new knowledge. By dynamically reallocating resources, the model can adapt to new events while minimizing the impact on previously learned patterns.
Secondly, we introduced clustering-based experience replay, which selectively preserves a small representative portion of the past data to reinforce the model's existing knowledge. This strategy ensures that the model retains valuable past experiences without overwhelming its memory capacity.
The experimental results we obtained demonstrate the effectiveness of our proposed strategies in adapting the model to new events while mitigating the occurrence of catastrophic forgetting. By leveraging temporal regularization and clustering-based experience replay, our framework enables the model to maintain a balance between learning new knowledge and retaining past knowledge.
Additionally, we investigated the relationship between the amount of memory allocated to experience replay and the benefits derived from our clustering-based sampling strategy. This analysis provided valuable insights into the importance of memory allocation in preserving and leveraging valuable past experiences during the model update process.

Recently, we have also explored how we can train transformer models in CL settings. Transformers are much larger compared to prior models such as convolutional neural networks and using weight regularization or experience replay on the full model may not lead to desired results. 
To improve upon these methods,
we introduce a  Continual Learning of Few-Shot Learners (CLIF) that aims to tackle the challenges posed by CL and FSL within a unified framework using a base transformer \cite{jin2021learn}. CLIF is designed to address the learning process where a model sequentially learns from a diverse range of NLP tasks. This sequential learning process enables the model to accumulate knowledge and enhance its ability to generalize to new tasks, while also retaining high performance on previously learned tasks.
In our work, we investigate how the model's generalization ability is affected in a CL setup. We evaluated various CL algorithms and proposed a novel approach which is based on regularized adapter generation. Our approach introduces regularization techniques specifically tailored to adapter modules, which are utilized to adapt the base transformer to individual tasks. Adapters are a small number of weights which are plugged in between the frozen layers of the pre-trained transformer. Since we keep the transformer layers to be frozen, its generalization ability is also maintained while more tasks are learned. To address catastrophic forgetting, we benefit from the idea of hyper-network, where a hyper-network is trained to generate the adapters weights for each task using a few samples from that task. We then use weight consolidation only on the hyper-network to mitigate forgetting effects. Since the hyper-network is much smaller than the base transformer,  using weight consolidation is much more effective.
Through our research, we contribute to the understanding of CL in the context of few-shot learners  and show that a base transformer can be adapted to new tasks using only a few samples.  

CL has not been explored significantly in multimodal settings.
To address this limitation, we introduce CLiMB, a benchmark specifically designed to investigate the challenges of learning multimodal tasks in a CL setting \cite{srinivasan2022climb}. CLiMB aims to systematically evaluate how upstream CL approaches can rapidly generalize to both new multimodal and unimodal tasks. Within CLiMB, we provide implementations of various CL algorithms, as well as a modified Vision-Language Transformer (ViLT) model that can be deployed for both multimodal and unimodal tasks.
Through our research using CLiMB, we discovered that common CL methods can indeed assist in mitigating forgetting during the learning of multimodal tasks. However, these methods fall short when it comes to enabling effective knowledge transfer across different tasks. Therefore, there is a need for the development of new CL algorithms specifically tailored to address the challenges of   multimodal settings.

If we train separate Adapter modules for each new task to address CL using transformers, forgetting effects will be minimal but we will overlook   the potential for cross-task knowledge transfer. To overcome this limitation, we proposed a   continual learning algorithm called Improvise to Initialize (I2I) \cite{srinivasan2023i2i}.
The key idea behind I2I is to initialize the adapters for incoming tasks by distilling knowledge from previously learned tasks' adapters. By leveraging the knowledge captured in the existing Adapters, I2I enables efficient knowledge transfer between tasks and avoids redundant training of separate Adapters for each new task.
To evaluate the effectiveness of I2I, we conducted experiments on our CLiMB benchmark.  By comparing the performance of adapters trained with I2I to those trained independently, we demonstrated that our algorithm consistently achieves better task accuracy. This improvement highlights the efficacy of our approach in facilitating knowledge transfer between task adapters. We also demonstrated that our approach is   efficient in terms of the size of the used adapter weights compared to the SOTA alternative.

We have also  proposed a novel CL architecture based on transformers that specifically targets bimodal vision-and-language tasks based on the idea of dynamic model expansion \cite{cai2023task}. Our approach involves dynamically increasing the number of learnable parameters and utilizing knowledge distillation to facilitate knowledge transfer. The additional parameters are utilized to specialize the network for each specific task, allowing for the sharing of information between tasks while mitigating the challenge of catastrophic forgetting.
One of the advantages of our approach is its scalability to a large number of tasks, as it requires minimal memory and time overhead. This scalability makes our model suitable for handling diverse and complex vision-and-language tasks. Our model achieves SOTA on challenging vision-and-language benchmarks, demonstrating its effectiveness in this domain.

There are a few works that try to extend CL to multiagent settings. In my research, I   addressed this setting by defining a CL setting over a network of collaborating networks \cite{rostami2017transfer,rostami2018multi}  which use distributed learning \cite{hao2016testing} to learn collectively from a sequence of tasks and improve their performances through sharing their high-level learned knowledge while preserving their local data privacy.

Finally, my recent research includes using diffusion models to implement the idea of generative replay to address catastrophic forgetting in CL settings \cite{hu2023encoding}. 

\section{Prospect for Future Research}

In much of my research, I have been primarily focused on utilizing abstract embedding spaces, particularly the last layer of a neural network before the softmax output layer, to facilitate knowledge transfer. It is   recognized that neural networks represent input data at various levels of abstraction within their hidden layers.
By organizing data in a hierarchical manner, hierarchical embeddings have the capacity to capture both high-level and low-level information. This hierarchical structure enables more effective processing of complex inputs. Consequently, it seems logical to leverage these hierarchical embedding spaces to facilitate knowledge transfer across diverse sets of tasks, depending on their relevance to each other.

Hierarchical embeddings have the ability to capture semantic relationships between entities. When data is organized hierarchically, similar entities are grouped together, and their embeddings reflect their proximity in the hierarchy. This mechanism allows for leveraging the inherent relationships and similarities among entities, which can lead to improved performance in tasks such as clustering and classification.
Furthermore, hierarchical embeddings facilitate generalization by encoding hierarchical similarities and differences. They are capable of capturing shared features at higher levels of the hierarchy, as well as specific characteristics at lower levels. This enables models to capture both global and local patterns, empowering them to generalize effectively to unseen tasks based on the level of similarity with prior seen tasks and handle variations within categories.

One significant advantage of hierarchical embeddings is their potential for model interpretability. The hierarchical structure enables intuitive navigation and understanding of the data. By examining embeddings at different levels of the hierarchy, valuable insights can be gained into the relationships and representations of different categories or concepts within the embedding space. This interpretability contributes to a better understanding of the underlying data.
Moving forward, I am highly interested in further exploring these directions in my future research. I aim to investigate how hierarchical embeddings can enhance knowledge transfer, improve performance in various tasks, facilitate generalization, and provide interpretability, ultimately advancing our understanding of complex data and models.

One of the key advantages of incorporating hierarchical embedding spaces into transfer learning is the ability to engage in reasoning based on representations within these latent spaces. In my past research, reasoning has been quite limited to tasks that share strong relationships such as sharing the same classes. In contrast, natural intelligence benefits from reasons in a much wider context across diverse tasks. Recent advancements have indicated that by combining symbolic logic with deep learning, we may be able to develop algorithms that facilitate reasoning in the context of neural networks. This fusion of symbolic logic and deep learning, known as neurosymbolic AI, has captured my keen interest, as I am eager to explore its potential for enabling transfer learning.
Prior to deep learning, symbolic logic provided us with a formal and structured means of representing knowledge. Through the use of logical symbols, predicates, and rules, we can encode information about relationships, concepts, and rules that are pertinent to a specific domain. This method of knowledge representation holds the promise of being shared and reused across various tasks, thereby facilitating transfer learning. However, a challenge arises in ensuring that the hidden layers of neural networks learn representations that possess these desirable properties. Part of the reason behind uninterpretability  of deep learning is that the reasoning mechanism of these models is quite unclear. It becomes imperative to explore how we can enforce the learning of representations in these hidden layers that align with symbolic logic.
By encoding specific rules or constraints within the embedding spaces, we can govern the relationships between entities or concepts. These encoded rules can then be transferred to related tasks, serving as a foundation for reasoning and decision-making in novel contexts. I   believe that this research direction remains relatively unexplored, presenting both a high-risk endeavor and the potential for substantial rewards in my future explorations.

\section{Conclusion}
I surveyed my works on various aspects of transfer learning within the realm of embedding spaces. My research delves into diverse learning settings, such as zero-shot learning, lifelong learning, and domain adaptation.
  I discussed the limitations of current methods and highlighted the need for innovative approaches that can effectively enable knowledge transfer. Additionally, I identified potential research directions that can be pursued in the near future to enhance the performance and generalizability of ML algorithms using transfer learning through embedding spaces.
In conclusion, by identifying the existing challenges in these areas and proposing potential research directions, I aimed to contribute to the advancement of transfer learning techniques in both the short-term and mid-term.

 \small

\bibliographystyle{plainnat}
\bibliography{ref}

\end{document}